\documentclass[letterpaper]{article} % DO NOT CHANGE THIS
\usepackage{aaai24}  % DO NOT CHANGE THIS
\usepackage{times}  % DO NOT CHANGE THIS
\usepackage{helvet}  % DO NOT CHANGE THIS
\usepackage{courier}  % DO NOT CHANGE THIS
\usepackage[hyphens]{url}  % DO NOT CHANGE THIS
\usepackage{graphicx} % DO NOT CHANGE THIS
\usepackage{natbib}  % DO NOT CHANGE THIS AND DO NOT ADD ANY OPTIONS TO IT
\usepackage{caption} % DO NOT CHANGE THIS AND DO NOT ADD ANY OPTIONS TO IT
\usepackage{algorithm}
\usepackage{algorithmic}
\usepackage{amsfonts, amsmath,amssymb}
\DeclareMathOperator{\signum}{sgn}

\usepackage{newfloat}
\usepackage{listings}
\DeclareCaptionStyle{ruled}{labelfont=normalfont,labelsep=colon,strut=off} % DO NOT CHANGE THIS
\floatstyle{ruled}
\newfloat{listing}{tb}{lst}{}
\floatname{listing}{Listing}
\title{Graph-based Prediction and Planning Policy Network (GP3Net) for scalable self-driving in dynamic environments using Deep Reinforcement Learning}
\author{
    Jayabrata Chowdhury \textsuperscript{\rm 1} \equalcontrib,
    Venkataramanan Shivaraman \textsuperscript{\rm 2} \equalcontrib,
    Suresh Sundaram \textsuperscript{\rm 1},
    P B Sujit \textsuperscript{\rm 2}
}
\affiliations{
    \textsuperscript{\rm 1}Indian Institute of Science, Bengaluru \hspace{2mm} \textsuperscript{\rm 2}Indian Institute of Science Education and Research, Bhopal \\
    jayabratac@iisc.ac.in,
    vshivaraman18@gmail.com,
    vssuresh@iisc.ac.in,
    sujit@iiserb.ac.in
}

\usepackage{bibentry}
\begin{document}

\maketitle

\begin{abstract}
Recent advancements in motion planning for Autonomous Vehicles (AVs) show great promise in using expert driver behaviors in non-stationary driving environments. However, learning only through expert drivers needs more generalizability to recover from domain shifts and near-failure scenarios due to the dynamic behavior of traffic participants and weather conditions. A deep Graph-based Prediction and Planning Policy Network (GP3Net) framework is proposed for non-stationary environments that encodes the interactions between traffic participants with contextual information and provides a decision for safe maneuver for AV. A spatio-temporal graph models the interactions between traffic participants for predicting the future trajectories of those participants. The predicted trajectories are utilized to generate a future occupancy map around the AV with uncertainties embedded to anticipate the evolving non-stationary driving environments. Then the contextual information and future occupancy maps are input to the policy network of the GP3Net framework and trained using Proximal Policy Optimization (PPO) algorithm. The proposed GP3Net performance is evaluated on standard CARLA benchmarking scenarios with domain shifts of traffic patterns (urban, highway, and mixed). The results show that the GP3Net outperforms previous state-of-the-art imitation learning-based planning models for different towns. Further, in unseen new weather conditions, GP3Net completes the desired route with fewer traffic infractions. Finally, the results emphasize the advantage of including the prediction module to enhance safety measures in non-stationary environments.
\end{abstract}

\section{Introduction}

\begin{figure}[h!]
    \centering
    \includegraphics[width=3.2in, height=1.9in]{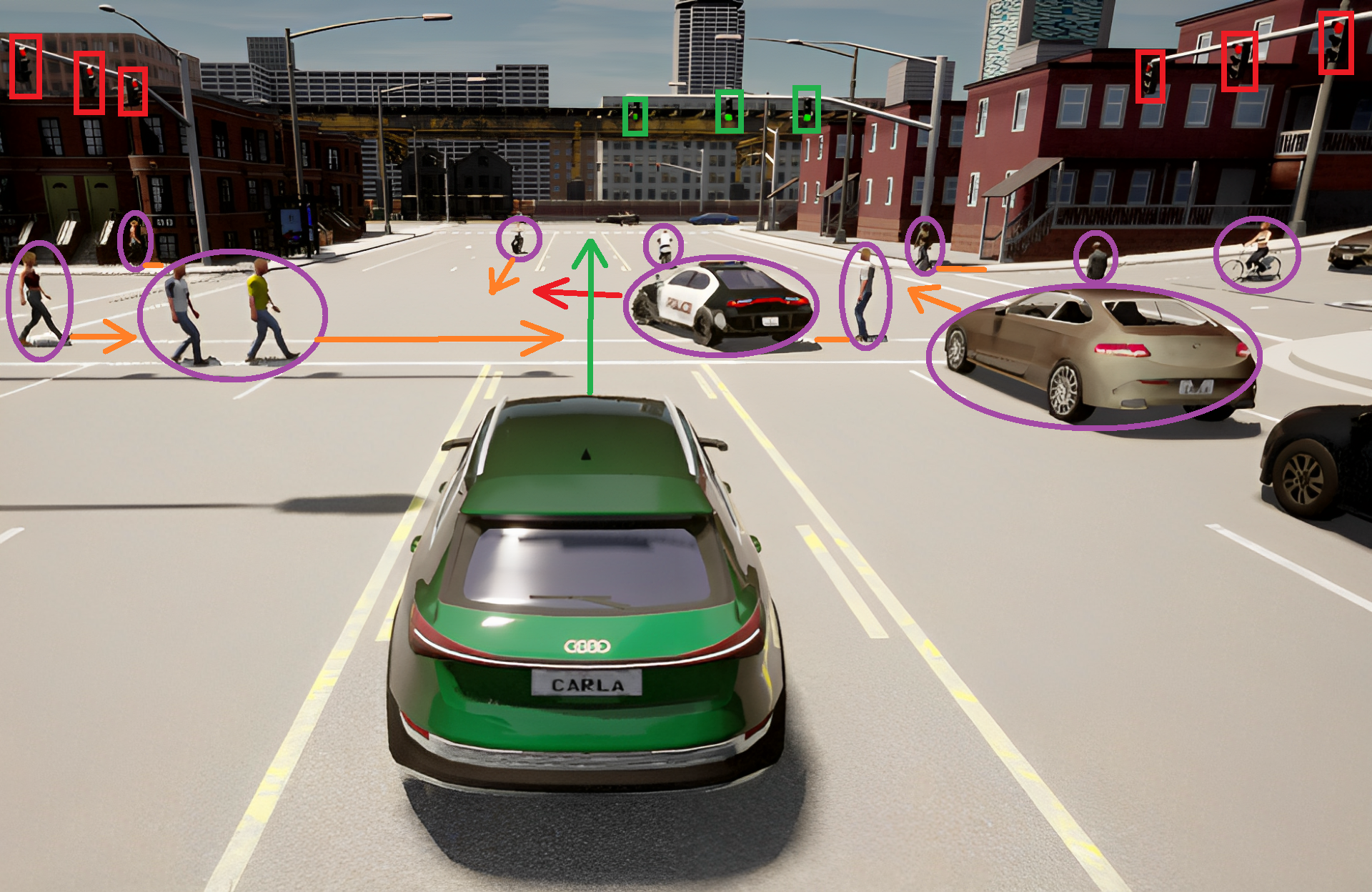}
    \caption{A typical intersection scenario with traffic lights, pedestrians, and vehicles. The scenario has pedestrians and vehicles moving on the desired path of AV. The figure has colour coded markers to show the context.}
    \label{intersection_region}
\end{figure}

Motion planning for AVs still has a long way to go due to the complexity of real-world driving environments, ensuring the safety and comfort of everyone. The AV must be able to carefully plan its maneuvers in urban and highway environments with varying traffic dynamics set by cars, bikers, pedestrians, and weather conditions. A typical intersection scenario in an urban environment is shown in Fig.\ref{intersection_region}.
Various traffic participants' different intentions and final goals (as shown by arrows in Fig.\ref{intersection_region}) make it challenging for AV to safely and comfortably move to its destination. Variable speeds of different traffic participants also make it difficult for AV to drive safely. A research finding on the United States \cite{choi2010crash} shows that wrong estimation of other traffic participants' speed is responsible for $8.4\%$ of critical reasons for crashes. Also, another problem of driving in an unstructured environment is that traffic pattern varies with the cooperative/noncooperative behaviors of other traffic participants. By utilizing communication channels as described in \cite{Bazzi2021AV_communication, Cui_2022_CVPR}, it becomes possible to determine the intentions and objectives of fellow traffic participants. This information simplifies the process for AVs to make informed decisions regarding their actions. However, vehicles use different communication protocols, and ensuring reliable real-time communications between all traffic participants is challenging. Hence, another way is to model the interactions of traffic participants through time and predict their future trajectories to understand their behaviors. An AV can make a safe and efficient decision in a non-stationary environment if it can model the interactions better and predict the future trajectories of other surrounding vehicles. Depending on the future trajectories of others', an AV can decide on a safe planning maneuver. Hence, there is a need to develop a motion planning algorithm that can handle non-stationary environments using the intention understanding from the predicted trajectories of other traffic participants.

Existing state-of-the-art methods can be broadly classified into rule-based, imitation learning-based, and reinforcement learning-based works. Recently \cite{Ville_Kyrki2021Rule_based} uses a rule-based decision-making system for navigating road intersections. However, due to the non-stationary nature of the environment, rules vary for each situation, and attempting to create generalized rules across diverse environments suffers from scalability issues. On the other hand, some models mimicking expert drivers, called Imitation Learning (IL) based motion planner, has been proposed in \cite{bansal2018chauffeurnet, rhinehart2020deep_imitative_model, Codevilla2018end, Cai2020Multimodal, Cai2021DiGNet}. The driving model in \cite{bansal2018chauffeurnet} learns driving decisions from expert human driver demonstrations. The work in \cite{rhinehart2020deep_imitative_model} uses expert driving data collected from CARLA \cite{dosovitskiy2017carla} simulator to learn expert behaviors. An end-to-end motion planning algorithm has been learned with IL in \cite{Cai2020Multimodal, Codevilla2018end}. However, the IL method suffers from a data distribution shift from training scenarios to unseen evaluation scenarios as \cite{bansal2018chauffeurnet} did not perform well in unknown highly interactive lane change scenarios. In \cite{Cai2021DiGNet}, a graph-based attention network has been used to model participant interactions, and a driver model learns to imitate the expert with IL. In another work \cite{Teng2023HIIL}, a Bird's Eye View (BEV) representation has been generated from the front camera view to use with IL for decision-making. The expert bias and distribution shift problems heavily influence the aforementioned IL-based works' performance. Since there are no near-collision scenarios in expert driving, it will be difficult for IL-based methods to perform in the real environment. 

Another popular approach to overcome the expert-based IL method is Reinforcement Learning (RL). In RL, the agent learns through their experience of completing the task. Recently works \cite{chen2019model, chen2021interpretable, ye2020automated, tang2022highway, Alizadeh2019LaneChangeHighway, Kaleb2021HDQN} employed RL for motion planning and maneuver decision-making for autonomous driving. In \cite{chen2019model, chen2021interpretable}, a BEV image-based representation has been used to represent the AV's surrounding environment. An RL agent learns to make maneuver decisions with this input representation. The works in \cite{ye2020automated, tang2022highway} learn lane-changing maneuvers in simulated handcrafted driving scenarios through policy optimization methods (\cite{schulman2017PPO, Harnoja2018SAC}). Recently, a work  \cite{chen2020conditional} used conditional Deep Q Network (CDQN) for stable driving decisions in the CARLA environment. However, this work did not include other vehicles, pedestrians, or traffic lights to model spatio-temporal interactions with AV. The works mentioned above need to recognize the intent of surrounding traffic participants. In \cite{huang2019online}, the intent of surrounding vehicles is calculated probabilistically to solve motion planning problems. However, the intent is modeled for a fixed number of vehicles without pedestrians and evaluated in a handcrafted driving scenario. The intentions of varying numbers of surrounding vehicles should be inferred to understand how the non-stationary environment can evolve. Hence, a predictive motion planning algorithm is required to model the spatio-temporal interactions for trajectory prediction and make a safe and efficient maneuver decision.  

The proposed deep Graph-based Prediction and Planning Policy Network (GP3Net) framework finds a way to handle distribution shifts, model the interactions to predict trajectories and plan safe maneuvers. The GP3Net's benefits are as follows:
\begin{enumerate}
    \item A deep spatio-temporal graphical model encodes interaction to depict the behaviors of traffic participants with AV. These modeled interactions are passed through the trajectory prediction module to provide predicted future trajectories with associated prediction uncertainties.
    \item The future mask generation module generates a BEV occupancy map with the prediction uncertainties, capturing the intentions of other traffic participants and probable evolution of a non-stationary environment in the future. 
    \item The AV neural architecture (state-encoder plus policy network) encodes past BEV masks and predicted future BEV masks together to give the final state.
    \item The AV neural architecture outputs the values of acceleration and steering directly. The whole AV neural architecture is updated end-to-end with the RL algorithm.
    \item The learned GP3Net driving agent outperforms the previous model DiGNet \cite{Cai2021DiGNet} and HIIL \cite{Teng2023HIIL} in CARLA Leaderboard and No-Crash motion planning benchmark in different towns and weathers. A Qualitative analysis highlights the importance of the graph-based trajectory prediction module.
\end{enumerate}

\section{Related Work}
\subsection{Trajectory prediction}
Trajectory prediction is a crucial part of autonomous driving involving spatiotemporal interactions between traffic participants. In \cite{ma2019trafficpredict, xu2020cf}, Long-Short Term Memory (LSTM) network has been used for modeling. However, these works need to model the influence of one traffic participant on another. Recently, \cite{shi2022social} proposes a tree-based attention method. Also, in \cite{li2021spatial}, spatio-temporal modeling has been by a graphical model. However, these approaches needed to provide uncertainties in trajectory prediction. In \cite{ivanovic2019trajectron}, a spatio-temporal dynamic graph models interactions, and LSTM networks model the state encoding of graph nodes. A Conditional Variational Auto-Encoder (C-VAE) learns the trajectory distribution. Samples from this probability distribution are taken for Gaussian Mixture Model (GMM)-based multimodal trajectory prediction. Trajectron predicts future trajectories with prediction uncertainties. GP3Net uses Trajectron architecture for trajectory prediction of surrounding traffic participants.

\subsection{End-to-end and semantic-based motion planning}
Many works directly use sensors (camera, LIDAR) to understand the surrounding environment and make decisions. In \cite{Codevilla2018end}, a directional high-level command is used to learn low-level steering and throttle values. The following work \cite{codevilla2019exploring} showed the limitations of IL for autonomous driving. Recently \cite{Teng2023HIIL} uses a front camera as a vision sensor and converts it to BEV representation of the driving scene. However, this work does not model rear-end side of the AV in the BEV. The rear-end side is crucial for AV because $64.2 \%$ of AV crashes happen from the rear end as reported in \cite{petrovic2020traffic}. GP3Net uses the semantic BEV representation for deciding safe maneuvers.
\begin{figure*}[t]
    \centering
    \includegraphics[scale=0.4]{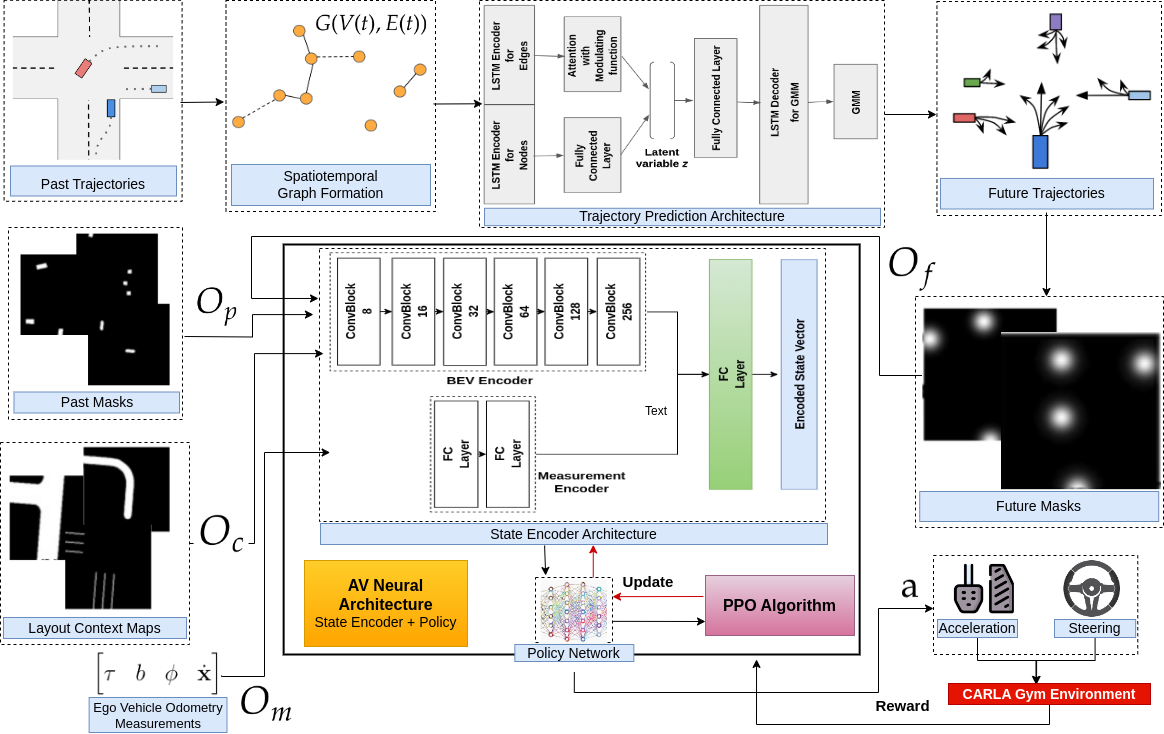}
    \caption{Schematic diagram of the GP3Net framework. The diagram shows the pipeline from input observations $\textbf{O}$ to control outputs $a$ and $\omega$, highlighting the future occupancy map generation by the trajectory prediction module, the state-encoder module, the PPO algorithm that optimizes the AV's state-encoder and policy network.
}
    \label{fig:GP3Net_Framework}
\end{figure*}

\section{The GP3Net Framework}
This section introduces the GP3Net motion planning framework that incorporates future trajectory predictions generated by a trajectory prediction module into Deep Reinforcement Learning-based motion planning. The subsections describe the self-driving task as a Partially Observed Markov Decision Process (POMDP). The primary input for the GP3Net consists of past and predicted future BEV occupancy maps containing information about road layout, vehicles, pedestrians, and traffic lights. The predicted future occupancy maps of vehicles are generated using a spatio-temporal graph-based trajectory prediction module based on \cite{ivanovic2019trajectron}. Fig. \ref{fig:GP3Net_Framework} is a flowchart showing the pipeline between input processing, future BEV occupancy map generation, and output control actions.

\subsection{POMDP Formulation} 
POMDP is defined for this task as a 6-tuple $(S, A, T, R, O, \gamma)$. Here, $S$ is a set of possible states in the environment, $A$ is the action space that the agent/AV can take in the environment, $T$ is the transition probability $T(s' |s, a)$ that represents the probability of AV moving to state $s'$ after taking action $a$ from state $s$. $R$ is the reward function $R: S \times A \rightarrow \mathbb{R}$ for the AV, $O$ is a set of observations the AV can get, and $\gamma \in (0, 1]$ is the discount factor. The goal is to learn a policy $\pi$ which can maximize the expected cumulative reward $\mathbb{E}[\sum_{t = 0}^{\infty} \gamma^{t} r_t]$, where $r_t$ is the reward obtained in the time step $t$. Following subsections define the observation space, action space, and reward function for the driving task in CARLA simulator.
\subsection{Input Preprocessing and Observation Space Design}
The observation space is $\textbf{O} = [O_{c}; O_{p}; O_{f}; O_{m}]$. This observation vector $\textbf{O}$ for RL-based AV consists of several components of particular categories. The observations $O_{c}, O_{p}$ and $O_{m}$, except $O_{f}$, are directly obtained from environment. The observations $O_{c}, O_{p}$ and $O_{f}$  are rasterized, segmented local BEV image/mask centered around the ego vehicle. The $O_{c}$ contains segmented BEV masks of the road layout, the mission route, and lane boundaries for contextual localization at time-step $t$. Observation $O_{p}$ contains past configurations masks of the surrounding vehicles, pedestrians, traffic lights, and stop signs for $k$ time-steps in the past, each $4$ time-steps apart. Observation $O_{m} = [\tau, b, \omega, \dot{\textbf{x}}]$ is a simple vector containing the AV's odometric measurements such as throttle $\tau$, braking $b$, steering $\omega$, and velocity $\dot{\textbf{x}} = (\dot{x}, \dot{y})$ at time-step $t$ as shown in Fig. \ref{fig:GP3Net_Framework}. The vector $O_{f}$ contains the predicted occupancy maps (added as masks) of surrounding vehicles for $l$ time-steps in the future, spaced $4$ time-steps apart. These future occupancy maps in $O_{f}$ are accurately generated by a spatio-temporal graph-based trajectory prediction module. This trajectory prediction model takes in the observed past \textit{trajectories} of surrounding vehicles/pedestrians and models their interactions as dynamic spatiotemporal graphs to accurately predict future trajectories. The future occupancy maps are generated using the future trajectories with respect to our AV.

\textbf{Graph-based Trajectory Prediction Module: }The trajectory prediction module takes past kinematics information of the surrounding vehicles as input to generate a distribution of future trajectories for all agents present in a scene. The most probable future trajectory is then selected for generating the future maps. This module doesn't assume a fixed number of vehicles in the sensor range of the AV. Let $\textbf{x}^{i}_{(t-k,\ t)} = [(x,y)_{t - k}, (x,y)_{t - (k - 1)},..., (x,y)_{t}]$ are the spatial coordinates of agent $i$ from timestep $t-k$ to $t$. The objective is first to predict the future trajectories $\textbf{x}^{1}_{(t+1,t+T)}, \textbf{x}^{2}_{(t+1,t+T)},..., \textbf{x}^{N}_{(t+1,t+T)}$ for $T$ timesteps for all $N$ agents which were present in the scene in that time window. Formally, the input for the model is $\textbf{x} = \left[\textbf{x}^{1,..,N}_{1:t_{obs}}, \dot{\textbf{x}}^{1,..,N}_{1:t_{obs}}, \ddot{\textbf{x}}^{1,..,N}_{1:t_{obs}} \right]$ and the output $\textbf{y}=\dot{\textbf{x}}^{1,..,N}_{(t_{obs}+1, t_{obs}+T)}$. The module contains a dynamic spatio-temporal graph model, deep generative CVAE with LSTM-based encoder/decoder, and GMM. This architecture directly models the velocity rather than the trajectories. When the future velocities are obtained, they are numerically integrated to produce future trajectories using single-integrator dynamics, as given below. Here, $\delta t$ is the difference of consecutive timesteps.
\begin{equation}
    \textbf{x}^{i}_{t+1} = \textbf{x}^{i}_t + \textbf{y}^{i}_t . \delta t
    \label{eq:single_integrator}
\end{equation}
The dynamic spatiotemporal graph representation effectively models dynamic interactions between the agents present in the scene. The graphs are created based on the spatial proximity of the agents.
The CVAE framework attempts to learn the probability distribution function $p(\textbf{y}|\textbf{x})$ as given below, where $z$ is a discrete latent variable. 
\begin{equation}
    p(\textbf{y}|\textbf{x}) = \sum_{z \in Z} p_{\psi}(\textbf{y}|\textbf{x}, z) p_{\phi}(z|\textbf{x})
    \label{eq:conditional_distribution}
\end{equation}
The encoder architecture consists of a node history encoder, a node future encoder (only during training), and an edge encoder. 
\begin{equation}
    e_{t, node}^{i} = LSTM(e_{t-1,node}^{i}, \textbf{x}^i_t; W_{NE})
    \label{eq:node_encoder}
\end{equation}
Both graph node history and future encoders have LSTM networks, as shown in Equ.\ref{eq:node_encoder} for encoding node data and temporal relations between them. 
\begin{equation}
    f_{t}^i = \left[ \textbf{x}_t^i ; \sum_{j \in N(i)} \textbf{x}_t^j  \right] ;  a_{t}^i = LSTM(a_{t-1}^i, f_{t}^i; W_{EE})
    \label{eq:edge_encoder}
\end{equation}
The edge encoder also uses LSTM network as given in Equ.\ref{eq:edge_encoder} and a modulating filter to encode the influence of other agents with smooth edge additions and removals if high-frequency changes occur in the graph. In Equ.\ref{eq:edge_encoder}, $[a;b]$ is concatenation, $N(i):$ set of neighbors of agent $i$. The latent variable $z$ is then sampled and fed to the decoder consisting of an LSTM network and GMMs. The outputs for the decoder's LSTM network are the GMMs' parameters. The GMMs are then used to produce the future velocities and trajectories. The future masks containing future occupancy maps for the surrounding vehicles are generated after obtaining the predicted future trajectories for all the vehicles inside AV's sensor range.

\textbf{Future Mask Generation:} After getting the predicted future trajectories $\textbf{x}^{1,...,N}_{t+1, t+T}$, the cartesian coordinates $\textbf{x}^{i}_{t+1, t+T}$ are converted to pixel coordinates $\textbf{p}^{i}_{t+1, t+T}$ centered around the AV for each vehicle/agent $i$. The existence of a vehicle in the future mask is denoted by a $2D$ symmetric Gaussian patch as shown in Fig.\ref{fig:GP3Net_Framework}. The pixel coordinates $\textbf{p}$ serve as the mean of the 2D symmetric Gaussian function. The symmetric Gaussian approach helps us account for the future prediction uncertainties of the vehicles, accounting for various factors such as sensor noise and occlusion. Equation of future vehicle patch is given by:
\begin{equation}
     g(x, y) = exp \left ({\frac{((x - x_{o})^2 + (y - y_{o})^2))}{2\sigma^{2}}} \right )
\end{equation}
In the above equation, $\sigma$ is the standard deviation of the Gaussian, $x_{o}$ and $y_{o}$ are the mean of the 2D Gaussian. For each timestep $t_{f}$ in future prediction horizon $T$ of the trajectory prediction model, a future mask is generated using pixel coordinates  $\textbf{p}^{1,...,N}_{t_{f}}$ of all $N$ vehicles in the method mentioned above. Therefore, $O_{f}$ consists of $l$ future masks. The observations are taken $4$ timesteps apart instead of $1$, and the inputs to the trajectory prediction module from timesteps $t-k$ to $t$ are passed in $4$ timestep intervals but processed as though inputs are of consecutive timesteps. This is further explained in the experiments section of the paper.

\textbf{Final State Encoder: }Once $O_{c}$, $O_{p}$, $O_{f}$, and $O_{m}$ are obtained, they are passed through a state-encoder (first part of AV Neural Architecture) which results in a column vector. The architecture for the state encoder, as shown in Fig. \ref{fig:GP3Net_Framework}, has two parts. The first part processes the BEV image masks $O_{c}, O_{p}$ and $O_{f}$ using a Convolutional Neural Network (CNN) as input channels and a simple Fully Connected (FC) layer. The measurement vector $O_{m}$ is encoded using two FC layers and concatenated with the output of the BEV encoder, passed through an FC layer, finally giving the final state representation for the RL algorithm.  

\subsection{Action Space}
The CARLA-Gym simulator takes three values for actions: throttle $\tau$, brake $b$, and steering $\omega$. However, to train the RL algorithm, action space is defined as a 2D continuous space $\textbf{a} = [\max\{\tau,b\} \cdot \signum(\tau - b), \ \omega]$, where sgn(.) is signum function. The first expression in $\textbf{a}$ is acceleration whose value ranges from -1 (maximum braking) to 1 (maximum throttle), and the steering value goes from -1 (maximum left steering) to 1 (maximum right steering). If the acceleration value is positive, it is passed to throttle $\tau$, and brake value $b$ is zero. If the acceleration value is negative, the magnitude of the acceleration is passed as brake $b$ value, and throttle $\tau$ is zero. To select actions from this continuous space, the policy network $\pi_{\theta}$ is trained to output parameters $\alpha$ and $\beta$ for a Beta distribution $\mathcal{B}(\alpha, \beta)$ for an improved policy gradient method \cite{chou2017improving}. The throttle and steering actions are sampled from this Beta distribution. This work uses separate Beta distributions for throttle and steering.

\subsection{Reward Structure}
The reward structure of RL is essential to train agents/AV to perform specific tasks. The reward is a superposition of different components, as given below.
\begin{equation}
    R = r_{route} + r_{halt} + r_{vel} + r_{pos} + r_{hd} + r_{act}
    \label{eq:reward_structure}
\end{equation}
where the AV is penalized, and the episode is terminated whenever the vehicle collides with anything, runs a traffic light or a stop sign, routes deviation, or blocks. $r_{route} = -1$ if route deviation $\delta \geq 3.5 m$; $r_{halt}=-1$ if AV's velocity $v \leq 0.1 m/s$; $r_{vel}=1-|v-v_{assigned}|/ v_{max}$ where $v_{max}$ is maximum velocity allowed and $v_{assigned}$ is assigned by the simulator; $r_{pos}=-0.5x$ where $x$ is the distance from the AV’s center and the center of the desired route; $r_{hd}= -\Delta_{h}$ absolute difference between heading of AV and route; $r_{act}=-0.1$ if $|\omega_{t} - \omega_{t-1}| > 0.01$.

\subsection{PPO Algorithm}
The AV's policy network $\pi$ (second part of AV neural architecture) learns using a state-of-the-art policy gradient-based RL algorithm called Proximal Policy Optimization (PPO) \cite{schulman2017PPO}. The usual form of loss function used in policy gradient methods is shown below. 

\begin{equation}
    \label{eq:policygradloss}
    L_{pg} = \mathbb{E}_{\pi} \left[ \hat{A}_t \log \pi(a_t \,| \,s_t, \theta) \right]
\end{equation}

Here, $\pi_{\theta}(a_t \,| \,s_t, \theta)$ is a stochastic policy with $\theta$ as parameter. $\hat{A}_t$ is an estimator of the advantage function at time $t$, and $\mathbb{E}_{\pi}$ represents the empirical average of a finite batch of samples.

However, updating the policy network $\pi_{\theta}$ for optimizing the objective loss function shown in Equ.\ref{eq:PPO} can lead to large policy updates, often resulting in instability and bad policy updates. A new objective function called the clipped surrogate objective, which places a constraint on policy updates, is proposed in PPO paper. This makes the training process much more stable and reliable when updated over multiple epochs of gradient ascent. Also, a Generalized Advantage Estimate (GAE) and adding an entropy bonus with their new clipped surrogate objective function is used, as shown in the following equation for PPO training of the GP3Net model.
\begin{equation}
\begin{aligned}
    \mathcal{L}_{ppo}=\mathbb{E} \Big[ \min (\tilde{r}(\theta)\hat{A}, \mbox{clip}(\tilde{r}(\theta), 1 - \epsilon, 1 + \epsilon)\hat{A}), \\ -c_{1}(V_{\phi}(s, a) - V_{targ}) 
    - c_{2}H(s, \pi_{\theta}) \Big]     
    \label{eq:PPO}
\end{aligned}
\end{equation}

Here in Equ.\ref{eq:PPO}, $\epsilon, c_{1} \mbox{ and } c_{2}$ are tunable hyper-parameters and $H(s, \pi_{\theta})$ denotes the entropy bonus. The generalized advantage estimate makes use of a learned state function $V_{\phi}(s, a)$ for computing the advantage function $\hat{A}$. The entropy bonus encourages the AV to explore adequately. Most importantly, during update, the whole AV Neural Architecture gets updated.

\section{Experiments}
\subsection{Training Setup}
This subsection describes the trajectory prediction module's training setup and the RL-based motion planning training.
\subsubsection{Trajectory Prediction Module Training}
First, trajectory data of surrounding vehicles are collected as observed by the CARLA simulator's \textit{roaming agent} in different towns for training the trajectory prediction module. The trajectory data contains rectangular coordinates $ \textbf{x}_{t} = (x_{t}, y_{t})$ of all vehicles at time-steps $t, t + 4\delta t, t + 8\delta t ...$ with each time-step spaced $\delta t=0.1$ seconds apart in simulation time. The trajectory prediction module is provided with 3.2 seconds of observed trajectories as input, divided into eight discrete time-steps $(t - 29\delta t, t - 25\delta t,...,t - \delta t)$, and the model predicts the trajectories up to 2.8 seconds, divided into seven time-steps $(t + 4\delta t, t + 8\delta t, ..., t + 28\delta t)$. 

The training starts by initializing the weights and setting the batch size to $64$ and the learning rate to $0.001$ with a decay rate of $0.9999$ to improve convergence. Adam optimizer updates the weight parameters of the model. Trajectory prediction models are typically benchmarked by comparing their performance against actual or simulated ground truth trajectories to evaluate accuracy and reliability. Here Mean Squared Error (MSE) metric evaluates the performance and accuracy of the model. The MSE is the mean $L_2$ error between the ground truth trajectories and predicted trajectories. The model was trained over 2000 steps and the MSE was evaluated intermittently over a validation set as shown in \cite{ivanovic2019trajectron}.
\subsection{RL Simulation Environment Setup}
The CARLA simulator is employed as a gym environment for training the RL algorithm and evaluating proposed RL-based motion planning. The training was conducted in parallel on six distinct town maps, namely Town 1 to Town 6, employing a vectorized environment configuration. This training approach on multiple maps simultaneously enhanced the replay buffer's diversity and improved the training process's efficiency. During the rollout phase, the AV is provided with a mission route at the beginning of each episode. This route is generated by randomly selecting a source and destination point and connecting them using a path search algorithm like $A^*$. The AV aims to successfully navigate the environment map and complete the assigned mission route. The episode would terminate if any of the following conditions were met: $1)$ collision with a vehicle, pedestrian, or obstacle in the map layout; $2)$ deviation from the given route; $3)$ traffic stagnation/halt and $4)$ violation of traffic lights and stop signs. Upon reaching the destination point, the AV was assigned a new mission route, continuing the rollout phase. The training and experiments were run in parallel on a Ubuntu 20.04 machine with 64 GB RAM and used two Nvidia 2080Ti GPU.
\subsubsection{RL-based Motion Planning Training}
The RL-based motion planning model is trained using the CARLA simulator \cite{dosovitskiy2017carla} as an RL gym environment. The environment setup is vectorized, and the model trains by running the RL training simulation on different town maps in parallel. The training process consists of two phases: (a) the rollout phase and (b) the training phase. In the rollout phase, the AV keeps interacting with the environment for several episodes, trying to navigate around the environment and reach its destination. At each step, the AV collects the tuple $(\textbf{O}, \textbf{a}, r, \textbf{O}^{'}, d)$ and stores it in its replay buffer. Here, $\textbf{O}$ and $\textbf{O}^{'}$ are observations of timestep $t$ and next timestep $t^{'}$, $a$ is action, $r$ is reward and $d$ is done/termination condition. The AV mainly collects experience from various scenarios offered by different town maps. The AV remains in this phase until it has accumulated sufficient experience in the replay buffer for moving into the training phase. The training phase mainly involves the AV updating its policy network $\pi_{\theta}$ based on the experience it accumulated in the rollout phase. The policy $\pi_{\theta}$ is updated based on the loss function Equ.\ref{eq:PPO}, and the AV again reverts to the rollout phase. After a few rounds, the AV's performance is evaluated, and several metrics are recorded to ensure that the AV is learning correctly and troubleshooting issues with proper tuning.

\section{Performance Evaluation}
\subsection{Evaluation Environment Setup}
To ensure a comprehensive evaluation of GP3Net, the performance benchmarking environment is setup based on the Leaderboard benchmarking suite described in the literature \cite{dosovitskiy2017carla} and NoCrash environment \cite{codevilla2019exploring}. These benchmarking suites offer diverse, complex scenarios across several towns, encompassing various weather conditions, numerical configurations, and behavior of vehicles and pedestrians. GP3Net's performance is assessed and recorded after conducting several episodes involving different environment configurations to ensure that the model can generalize well in complex unseen scenarios.

\subsection{Performance Evaluation Metrics}
\begin{table*}[h]
\centering
\begin{tabular}{|c|cc|cc|cc|cc|cc|cc|}
\hline
 &
  \multicolumn{2}{c|}{\begin{tabular}[c]{@{}c@{}}Town 03\\ urban\end{tabular}} &
  \multicolumn{2}{c|}{\begin{tabular}[c]{@{}c@{}}Town 04\\ mixed\end{tabular}} &
  \multicolumn{2}{c|}{\begin{tabular}[c]{@{}c@{}}Town 05\\ urban\end{tabular}} &
  \multicolumn{2}{c|}{\begin{tabular}[c]{@{}c@{}}Town 06\\ highway\end{tabular}} &
  \multicolumn{2}{c|}{\begin{tabular}[c]{@{}c@{}}Town 01\\ urban\end{tabular}} &
  \multicolumn{2}{c|}{\begin{tabular}[c]{@{}c@{}}Town 02\\ urban\end{tabular}} \\ \hline
Models &
  \multicolumn{1}{c|}{SR $\uparrow$} &
  DS $\uparrow$ &
  \multicolumn{1}{c|}{SR $\uparrow$} &
  DS $\uparrow$ &
  \multicolumn{1}{c|}{SR $\uparrow$} &
  DS $\uparrow$ &
  \multicolumn{1}{c|}{SR $\uparrow$} &
  DS $\uparrow$ &
  \multicolumn{1}{c|}{SR $\uparrow$} &
  DS $\uparrow$ &
  \multicolumn{1}{c|}{SR $\uparrow$} &
  DS $\uparrow$ \\ \hline
CILRS &
  \multicolumn{1}{c|}{43.2} &
  0.56 &
  \multicolumn{1}{c|}{16.2} &
  0.28 &
  \multicolumn{1}{c|}{38.2} &
  0.52 &
  \multicolumn{1}{c|}{35.2} &
  0.48 &
  \multicolumn{1}{c|}{42.3} &
  - &
  \multicolumn{1}{c|}{24.7} &
  - \\ \hline
MLP &
  \multicolumn{1}{c|}{62.3} &
  0.72 &
  \multicolumn{1}{c|}{60.5} &
  0.70 &
  \multicolumn{1}{c|}{60.0} &
  0.69 &
  \multicolumn{1}{c|}{72.2} &
  0.79 &
  \multicolumn{1}{c|}{-} &
  - &
  \multicolumn{1}{c|}{-} &
  - \\ \hline
GCN (U) &
  \multicolumn{1}{c|}{68.8} &
  0.77 &
  \multicolumn{1}{c|}{68.5} &
  0.76 &
  \multicolumn{1}{c|}{70.2} &
  0.78 &
  \multicolumn{1}{c|}{73.8} &
  0.80 &
  \multicolumn{1}{c|}{-} &
  - &
  \multicolumn{1}{c|}{-} &
  - \\ \hline
GCN (D) &
  \multicolumn{1}{c|}{74.2} &
  0.80 &
  \multicolumn{1}{c|}{64.0} &
  0.72 &
  \multicolumn{1}{c|}{71.5} &
  0.79 &
  \multicolumn{1}{c|}{74.8} &
  0.81 &
  \multicolumn{1}{c|}{-} &
  - &
  \multicolumn{1}{c|}{-} &
  - \\ \hline
GAT &
  \multicolumn{1}{c|}{76.0} &
  0.82 &
  \multicolumn{1}{c|}{71.0} &
  0.78 &
  \multicolumn{1}{c|}{81.5} &
  0.87 &
  \multicolumn{1}{c|}{78.2} &
  0.83 &
  \multicolumn{1}{c|}{-} &
  - &
  \multicolumn{1}{c|}{-} &
  - \\ \hline
DiGNet (CTL) &
  \multicolumn{1}{c|}{72.8} &
  0.79 &
  \multicolumn{1}{c|}{67.5} &
  0.75 &
  \multicolumn{1}{c|}{69.8} &
  0.77 &
  \multicolumn{1}{c|}{73.2} &
  0.79 &
  \multicolumn{1}{c|}{-} &
  - &
  \multicolumn{1}{c|}{-} &
  - \\ \hline
DiGNet (Full) &
  \multicolumn{1}{c|}{80.2} &
  0.85 &
  \multicolumn{1}{c|}{67.5} &
  0.75 &
  \multicolumn{1}{c|}{69.8} &
  0.77 &
  \multicolumn{1}{c|}{78.2} &
  0.79 &
  \multicolumn{1}{c|}{-} &
  - &
  \multicolumn{1}{c|}{-} &
  - \\ \hline
\textbf{GP3Net (ours)} &
  \multicolumn{1}{c|}{\textbf{82.5}} &
  \textbf{0.87} &
  \multicolumn{1}{c|}{\textbf{75.0}} &
  \textbf{0.85} &
  \multicolumn{1}{c|}{\textbf{82.8}} &
  \textbf{0.91} &
  \multicolumn{1}{c|}{\textbf{82.5}} &
  \textbf{0.85} &
  \multicolumn{1}{c|}{\textbf{92.5}} &
  \textbf{0.96} &
  \multicolumn{1}{c|}{\textbf{93.1}} &
  \textbf{0.91} \\ \hline
\end{tabular}
\caption{A comparative result showing the mean success rate and the driving score of GP3Net and other SOTA works on different towns.}
\label{Comparison_table}
\end{table*}
Specific performance evaluation metrics are employed to measure the performance of the predictive reinforcement learning-based motion planning framework GP3Net by the benchmarking suites aforementioned in the previous subsection. These metrics include the Success Rate (SR) and Driving Score (DS). The SR is calculated as the percentage of instances where the AV reaches its intended destination without experiencing any collision. This metric directly assesses the model's ability to navigate and complete the designated mission routes safely and successfully. The DS is a composite metric that combines the AV's route completion success with penalties incurred due to collisions and traffic rule infractions. It is formulated as the product of the percentage of completed routes and a penalty factor. 

\subsection{Quantitative Benchmark Results}
The GP3Net's performance was evaluated in different towns and weather conditions for $100$ episodes with five different seeds. Here Table \ref{Comparison_table} and Table \ref{tab:vary_weather} display the mean comparative results with other previously proposed solutions, some of which use a BEV perspective for motion planning similar to GP3Net. It can be observed in Table \ref{Comparison_table} that the predictive RL-based motion planner, GP3Net, outperforms DiGNet \cite{Cai2021DiGNet}, its variations, and other works like CILRS \cite{codevilla2019exploring} in terms of the success rate and the driving score in several towns. The mean improvement in the success rate and driving score is $3.85\%$ and $8\%$, respectively. The standard deviation is $1.5 \%$. GP3Net has a large improvement in Town 4 and Town 6 which consists of both mixed and highway environments, showing capability to handle well in higher speeds. This generalization to a wide range of scenarios can be attributed to the GP3Net's understanding of how a non-stationary environment can evolve with the trajectory prediction module. 

% \begin{figure}[h]
%     \centering
%     \begin{subfigure}
%     \centering     \includegraphics[scale=0.35]{Images/qual_result_reward.png}     \label{fig:qual_rew_analysis}
%     \end{subfigure}
%     % \includegraphics[scale=0.32]{Images/qual_result_reward.png}
%     % \caption{Plots of reward obtained with no. of steps with and without trajectory prediction module}
%     % \label{fig:qual_rew_analysis}
% % \end{figure}
%     \begin{subfigure}
%     \centering
%         \includegraphics[scale=0.35]{Images/qual_result_collision.png}
%         \label{fig:qual_collision_analysis}
%     \end{subfigure}
%     \caption{Plots of (a) reward obtained and (b) vehicle collisions with and without trajectory prediction module.}
%     %\label{fig:qual_collision_analysis}
% \end{figure}

\begin{figure}[h]
  \centering
  \begin{tabular}{@{}c@{}}
    \includegraphics[scale=0.35]{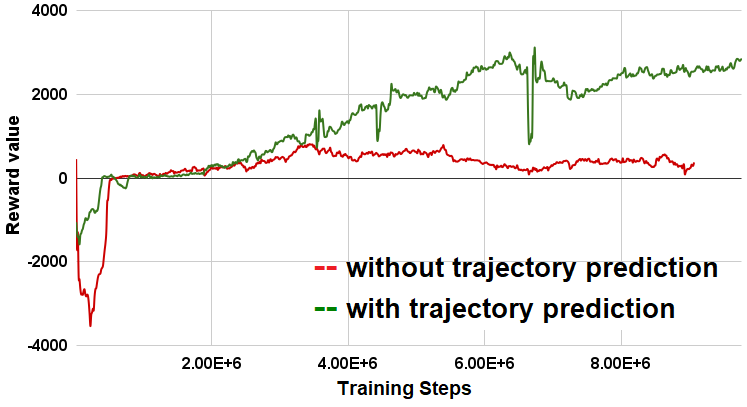} %\\[\abovecaptionskip]
    %\small (a) An image
  \end{tabular}

  % \vspace{\floatsep}

  \begin{tabular}{@{}c@{}}
    \includegraphics[scale=0.35]{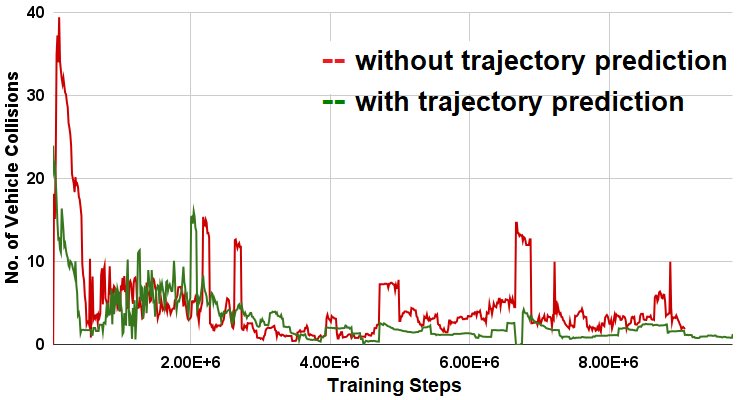}% \\[\abovecaptionskip]
    %\small (b) Another image
  \end{tabular}
    \caption{Plots of (a) reward obtained and (b) vehicle collisions with and without trajectory prediction module.}
    \label{fig:qual_analysis}
\end{figure}

\begin{table}[h]
\centering
\begin{tabular}{|l|c|c|c|c|}
\hline
\textbf{\begin{tabular}[c]{@{}l@{}}Weather \\ Condition\end{tabular}} &
  \textbf{\begin{tabular}[c]{@{}c@{}}T1 \\ (ours)\end{tabular}} &
  \textbf{\begin{tabular}[c]{@{}c@{}}T2 \\ (ours)\end{tabular}} &
  \textbf{\begin{tabular}[c]{@{}c@{}}T1 \\ (HIIL)\end{tabular}} &
  \textbf{\begin{tabular}[c]{@{}c@{}}T2 \\ (HIIL)\end{tabular}} \\ \hline
Wetnoon        & \textbf{96}   & \textbf{95.9} & 84 & 84 \\ \hline
ClearSunset    & \textbf{97.3} & \textbf{92}   & 88 & 88 \\ \hline
ClearNoon      & \textbf{92}   & \textbf{94}   & 89 & 89 \\ \hline
HardRainNoon   &\textbf{96}   & \textbf{88}   & 80 & 80 \\ \hline
SoftRainSunset & \textbf{93.3} & \textbf{94}   & 86 & 86 \\ \hline
WetSunset      & \textbf{98 }  & \textbf{94}   & 85 & 85 \\ \hline
WetCloudSunset & \textbf{98 }  & \textbf{92}  & 84 & 84 \\ \hline
SoftRainNoon   & \textbf{88}   & \textbf{91.4} & \textbf{88} & 88 \\ \hline
\end{tabular}
\caption{Success-rate Benchmarking results on T1: Town 1 and T2: Town 2 in different weather conditions.}
\label{tab:vary_weather}
\end{table}
Another experiment was performed to test the resilience of GP3Net to different weather conditions, whose results are shown in Table \ref{tab:vary_weather}. The test was conducted in Towns 1 and 2 under eight weather conditions mentioned in Table \ref{tab:vary_weather}. According to Table \ref{tab:vary_weather}, GP3Net is robust to rainy weather conditions and does a better job of navigating the environment successfully compared to another work, HIIL \cite{Teng2023HIIL}.
The insights about the future trajectories provided by the proposed trajectory prediction module have positively impacted the RL-based motion planner's decision-making during training and evaluation time, making it safe and cautious of the non-stationary environment. The following section discusses the effectiveness of our graph-based prediction module.

\subsection{Qualitative Analysis}
During training experiments, we plotted quantities such as total reward and the number of vehicle collisions for monitoring the training process for two
cases: 1) With the trajectory prediction module and 2) without the trajectory prediction module. Fig.\ref{fig:qual_analysis} shows a significant difference in the training plots. After the $5$ M steps, both the models have learned correctly not to deviate from the route or to stop when the lights turn red. However, the models were yet to learn to avoid collisions with other vehicles. The model with a trajectory prediction module showed a better capability to avoid collisions as the training progressed compared to the one which did not use a trajectory prediction module.

\textbf{Ablation Study: }In addition to the leaderboard suite, GP3Net's performance was benchmarked on the NoCrash suite for analyzing the framework's behavior in dense environments. Several metrics are utilized, such as end reach \%, success rate \%, penalty factor, crash rates, vehicle halts, and traffic lights passed were recorded to perform this analysis. The quantitative evaluation was done in Towns 1 and 2 in training and testing weather conditions which are displayed in Table \ref{tab:ablation_results}. It can be observed that the success rate of GP3Net in Town 2, though competitive, is not as high as in Town 1 due to a much denser population of vehicles and walkers since Town 2 is smaller in size compared to Town 1. It is to be noted that the NoCrash suite has twice the walker population than that in the Leaderboard suite.
\begin{table}[h]
\centering
\begin{tabular}{|l|l|l|l|l|}
\hline
\textbf{\begin{tabular}[c]{@{}l@{}}Performance \\ Metric\end{tabular}} & \textbf{\begin{tabular}[c]{@{}l@{}}T1 \\ train\end{tabular}} & \textbf{\begin{tabular}[c]{@{}l@{}}T1 \\ new\end{tabular}} & \textbf{\begin{tabular}[c]{@{}l@{}}T2 \\ train\end{tabular}} & \textbf{\begin{tabular}[c]{@{}l@{}}T2 \\ new\end{tabular}} \\ \hline
End Reach (\%) & 100 & 100 & 96 & 90 \\ \hline
Succ. Rate (\%) & 96 & 98 & 94 & 80 \\ \hline
Driving Score & 0.976 & 0.982 & 0.976 & 0.918 \\ \hline
Penalty Factor & 0.976 & 0.982 & 0.992 & 0.941 \\ \hline
Score Route & 1 & 1 & 0.984 & 0.977 \\ \hline
Layout crash & 0 & 0 & 0 & 0.047 \\ \hline
Walker crash & 0.039 & 0 & 0 & 0.151 \\ \hline
Vehicle crash & 0.018 & 0.039 & 0.083 & 0.167 \\ \hline
Vehicle halt & 0 & 0 & 0.313 & 1.556 \\ \hline
Lights met & 4.4 & 4.4 & 3.7 & 3.7 \\ \hline
Lights passed & 4.38 & 4.36 & 3.7 & 3.7 \\ \hline
\end{tabular}
\caption{Ablation study on CARLA NoCrash benchmarking suite. T1: Town 1; T2: Town 2.}
\label{tab:ablation_results}
\end{table}

\section{Conclusion}
This paper discusses the challenges faced by rule-based and imitation learning methods that rely on expert demonstrations. It highlights the limited ability of these methods to recover from domain shifts and near-failure scenarios in non-stationary environments. To mitigate this problem, this paper presents a deep Graph-based Prediction and Planning Network (GP3Net) framework that encodes spatio-temporal dynamic interactions of different traffic participants using a graphical model to infer their future behaviors and combines the advantage of exploration in reinforcement learning. GP3Net improves AV performance compared to the previous graph-encoded imitation learning-based policy in CARLA benchmarking driving scenarios for urban and highway environments. The qualitative study also clearly indicates the importance of the proposed prediction module for safe motion planning in dynamic driving scenarios. Future work will look at other methods to model interactions with the dynamics of traffic participants involved and how they can influence the planning capabilities of AV.

\bibliography{aaai24}
\newpage
% \newpage
\section{Supplementary Materials for GP3Net}
The supplementary materials consists of 1) More qualitative analysis 2) More quantitative results for different towns, 3) Architectural details of the neural network models used in GP3Net, 4) Hyper-parameter settings for GP3Net and 5) video recording of trained GP3Net based AV in different towns in different weathers. 

\subsection{Driving Videos by GP3Net}
There are six videos included in the videos folder. The videos show many scenarios in different towns (small cities, big cities, highways, and mixed towns) and weather conditions (rainy and clear). There are urban intersections, junctions, highway lane merging, and diversions. The GP3Net performed better in all scenarios.

\subsection{Qualitative Analysis}
\begin{figure}[h]
    \centering
    \includegraphics[scale=0.43]{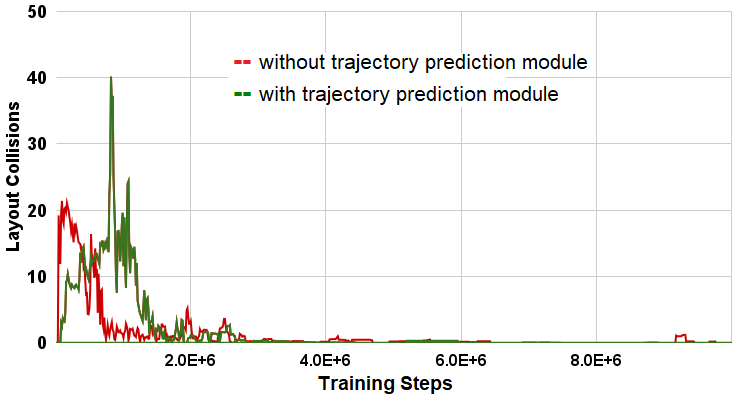}
    \caption{Plot of the layout collisions with and without trajectory prediction module.}
    \label{fig:layout_collision}
\end{figure}
\begin{figure}[h]
    \centering
    \includegraphics[scale=0.43]{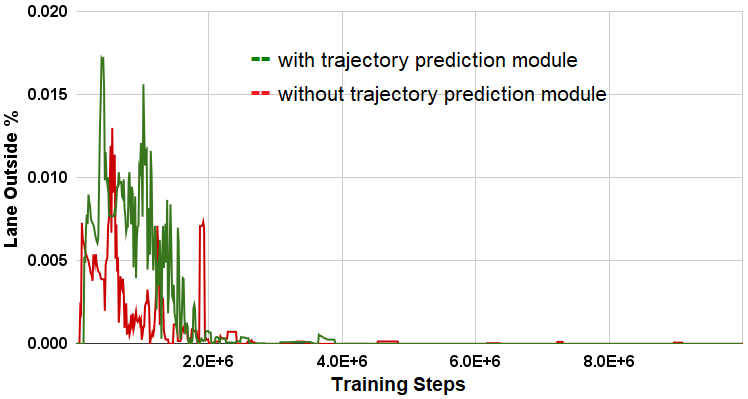}
    \caption{Plot of the percentage of cases going outside lane with and without trajectory prediction module.}
    \label{fig:lane_outside_per}
\end{figure}
This section shows some plots about the importance of the prediction module in motion planning for AVs. A PPO algorithm is trained without the prediction module to check its performance. The following plots show the performance in different metrics defined by CARLA with and without the trajectory prediction module.
\begin{figure}[H]
    \centering
    \includegraphics[scale=0.43]{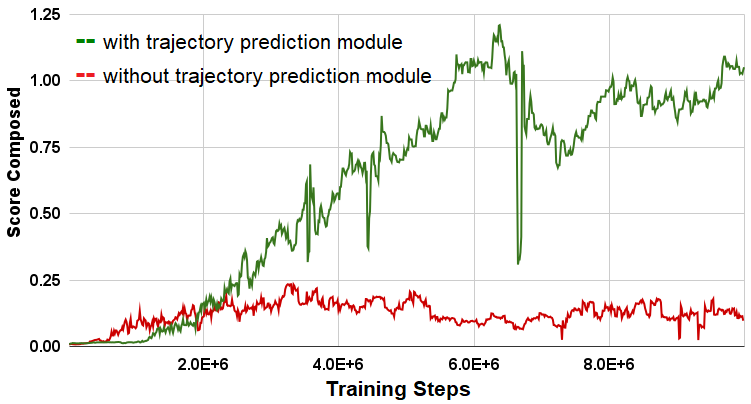}
    \caption{Plot of the score composed with and without trajectory prediction module.}
    \label{fig:score_composed}
\end{figure}
In Fig.\ref{fig:layout_collision} and Fig. \ref{fig:lane_outside_per}, it can be seen that up to 2M steps, the GP3Net model without trajectory prediction performed better. However, after 2M training steps, the GP3Net with trajectory prediction module started performing well.
\begin{table*}[h]
\centering
\begin{tabular}{|l|l|l|l|l|l|l|}
\hline
Performance Metric &
  \multicolumn{1}{c|}{\begin{tabular}[c]{@{}c@{}}Town01\\ Urban\end{tabular}} &
  \begin{tabular}[c]{@{}l@{}}Town02\\ Urban\end{tabular} &
  \begin{tabular}[c]{@{}l@{}}Town03\\ Urban\end{tabular} &
  \begin{tabular}[c]{@{}l@{}}Town04\\ Mixed\end{tabular} &
  \begin{tabular}[c]{@{}l@{}}Town05\\ Urban\end{tabular} &
  \begin{tabular}[c]{@{}l@{}}Town06\\ Highway\end{tabular} \\ \hline
End Reach (\%)    & 100    & 95.43  & 96.25  & 93.75  & 97.86  & 95.0   \\ \hline
Success Rate (\%) & 92.5   & 93.14  & 82.5   & 75.0   & 82.8   & 82.5   \\ \hline
Driving Score     & 0.96   & 0.91   & 0.87   & 0.85   & 0.91   & 0.91   \\ \hline
Penalty Factor    & 0.9588 & 0.9840 & 0.9062 & 0.8721 & 0.9223 & 0.914  \\ \hline
Score Route       & 1.0    & 0.9994 & 0.9866 & 0.9794 & 0.9916 & 0.9832 \\ \hline
Layout Crash      & 0      & 0      & 0      & 0.0115 & 0      & 0      \\ \hline
Walker Crash      & 0.0922 & 0      & 0.2141 & 0.0419 & 0.015  & 0.0104 \\ \hline
Vehicle Crash     & 0      & 0.0508 & 0.0320 & 0.0579 & 0.1291 & 0.0865 \\ \hline
Vehicle Halt      & 0      & 0.2509 & 0.0394 & 0.0439 & 0.0572 & 0.0296 \\ \hline
Lights met        & 4.7    & 3.76   & 7.638  & 8.65   & 8.2357 & 6.025  \\ \hline
Lights passed     & 4.675  & 3.7371 & 7.5375 & 8.5375 & 8.2214 & 5.975  \\ \hline
\end{tabular}
\caption{Detailed performance of GP3Net on Leaderboard Benchmarking scenarios in different towns}
\label{tab:Quant_results}
\end{table*}
In Fig.\ref{fig:score_composed}, the advantage of the trajectory prediction module is visible. After 2M training steps, there is improvement in the metric score composed. This metric combines infractions and penalty factors and provides the final score for driving.

\subsection{Detailed Quantitative Results}
The detailed quantitative results for the CARLA Leaderboard benchmarking scenarios are given in Table.\ref{tab:Quant_results}.

\subsection{Trajectory Prediction Architecture Details}
The trajectory prediction module uses a deep generative trajectory prediction architecture which is has Conditional Variational Auto Encoder (CVAE) to generate a distribution of potential trajectories. CVAE, a latent variable model, rely on Gaussian Mixture Models (GMM) to produce multimodal distribution outputs. The encoder and decoder of the CVAE consists of Long Short Term Memory (LSTM) networks for modeling temporal data such as trajectories and system evolution. After representing the nodes and edges as a vector, they are encoded using a Node Encoder and Edge Encoder respectively, whose hidden dimensions are shown in Table \ref{tab:traj_arch}. The encoded nodes and edges are sent to an attention module and low-pass filter $M$. The probability density function $p(z|\textbf{x})$ is captured by a Fully Connected Layer. Here, $z$ is the latent variable and $\textbf{x}$ is the input to the model. After sampling $z$ from $p(z|\textbf{x})$, it is passed through the decoder LSTM network which outputs the parameters for the GMM, from which multimodal trajectories are sampled. Table \ref{tab:traj_arch} shows the Trajectory Prediction Architecture details.

\begin{table}[h]
\centering
\begin{tabular}{|c|c|}
\hline
\textbf{Network Architecture Components}                                               & \textbf{Value} \\ \hline
Batch Size                     & 16      \\ \hline
Learning Rate                  & 0.001   \\ \hline
Minimum Learning Rate          & 0.00001 \\ \hline
Learning Decay Rate            & 0.9999  \\ \hline
Edge Addition distance (vehicles)         & 45 meters    \\ \hline
LSTM Encoder Edge              & 8       \\ \hline
\begin{tabular}[c]{@{}c@{}}LSTM Node History Encoder \\ Hidden Dimensions\end{tabular} & 32                  \\ \hline
\begin{tabular}[c]{@{}c@{}}LSTM Node Future Encoder\\ Hidden Dimensions\end{tabular}   & 32                  \\ \hline
FC Layer $p(z|\textbf{x})$ Dimensions   & 16      \\ \hline
LSTM Decoder Hidden Dimensions & 128     \\ \hline
GMM Components                 & 16      \\ \hline
Past Data Length               & 8 steps (3.2 sec) \\ \hline
Future Prediction Horizon      & 7 steps (2.8 sec) \\ \hline
\end{tabular}
\caption{Trajectory Prediction Module Architecture and Hyperparameters Details}
\label{tab:traj_arch}
\end{table}

\subsection{State Encoder Architecture Details}
The State Encoder consists of Convolutional Neural Network (CNN) for encoding the BEV (past and generated future) masks and a Multi-Layered Perceptron to encode the odometry values of AV. After concatenating the outputs from these two networks, it is further processed using a Fully-Connected Layer giving the final state encoding. The details of the State Encoder architecture is displayed in Table \ref{tab:StateEncArch}.

\begin{table}[H]
\centering
\begin{tabular}{|c|c|c|c|}
\hline
\textbf{\begin{tabular}[c]{@{}c@{}}State Encoder \\ Layers\end{tabular}} &
  \textbf{\begin{tabular}[c]{@{}c@{}}IN  \\ Channels/\\ Dims\end{tabular}} &
  \textbf{\begin{tabular}[c]{@{}c@{}}OUT \\ Channels/\\ Dims\end{tabular}} &
  \textbf{\begin{tabular}[c]{@{}c@{}}Filter\\ Size\end{tabular}} \\ \hline
Conv 1 (BEV)                                                         & 21  & 8   & 5 \\ \hline
Conv 2 (BEV)                                                         & 8   & 16  & 5 \\ \hline
Conv 3 (BEV)                                                         & 16  & 32  & 5 \\ \hline
Conv 4 (BEV)                                                         & 32  & 64  & 3 \\ \hline
Conv 5 (BEV)                                                         & 64  & 128 & 3 \\ \hline
Conv 6 (BEV)                                                         & 128 & 256 & 3 \\ \hline
\begin{tabular}[c]{@{}c@{}}Fully Connected\\ (Odometry)\end{tabular} & 256 & 256 & - \\ \hline
\begin{tabular}[c]{@{}c@{}}Fully Connected\\ (Odometry)\end{tabular} & 256 & 256 & - \\ \hline
Fully Connected                                                      & 256 & 512 & - \\ \hline
Fully Connected                                                      & 512 & 256 & - \\ \hline
\end{tabular}
\caption{State Encoder Architecture }
\label{tab:StateEncArch}
\end{table}

\subsection{PPO Policy Network Architecture Details and Hyperparameters}
The policy network $\pi$ like in many RL works is a Multi-Layered Perceptron (MLP). This policy network $\pi$ gives the values of $\alpha$ and $\beta$ parameters of Beta distribution of throttle and steering, resulting in two $\alpha$s and two $\beta$s. The details for policy network architecture and hyperparameters used for PPO training is shown in Table \ref{tab:policy_network}.
\begin{table}[H]
\centering
\begin{tabular}{|c|cc|}
\hline
\textbf{Policy Network Layers} & \multicolumn{1}{c|}{\textbf{IN dims}} & \textbf{OUT dims} \\ \hline
Fully Connected            & \multicolumn{1}{c|}{256} & 256 \\ \hline
Fully Connected ($\alpha$) & \multicolumn{1}{c|}{256} & 2   \\ \hline
Fully Connected ($\beta$)  & \multicolumn{1}{c|}{256} & 2   \\ \hline
\textbf{PPO Hyperparameters}   & \multicolumn{2}{c|}{\textbf{Values}}                      \\ \hline
Clip ratio $\epsilon$      & \multicolumn{2}{c|}{0.2}       \\ \hline
Epochs                     & \multicolumn{2}{c|}{30}        \\ \hline
Gamma $\gamma$             & \multicolumn{2}{c|}{0.99}      \\ \hline
Target KL                  & \multicolumn{2}{c|}{0.01}      \\ \hline
GAE $\lambda$              & \multicolumn{2}{c|}{0.97}      \\ \hline
Learning Rate              & \multicolumn{2}{c|}{0.00003}   \\ \hline
\end{tabular}
\caption{Policy Network Architecture and Hyperparameter Details}
\label{tab:policy_network}
\end{table}

\subsection{Data collection settings for trajectory prediction module}
The training data was collected using the CARLA simulator’s Autopilot/Roaming Agent in Town 1. For several episodes, this autopilot vehicle navigates from source to destination, recording the trajectories of all vehicles at each time step $t$. The trajectories are represented by a sequence of local rectangular coordinates $\textbf{x}_{t} = (x_{t}, y_{t})$ of the agents at time step $t$, as observed by the roaming agent. The dataset contains the timestamp of the recorded trajectory data, the index of the agent, and the trajectory coordinates. The data for vehicles and pedestrians are collected every 4-time steps (0.4 seconds) and stored in separate datasets, as we train two models for the former and the latter.

\end{document}